%% file: iclr2027_conference.tex
\documentclass{article} 
\usepackage{iclr2027_conference,times}

\input{math_commands.tex}

\usepackage{algorithm}
\usepackage{algpseudocode}
\usepackage{wrapfig}
\usepackage{hyperref}
\usepackage{url}
\usepackage{twemojis}
\usepackage{booktabs}
\usepackage{multirow}
\usepackage{colortbl}
\usepackage{graphicx}
\usepackage{xspace}

\definecolor{deltapos}{HTML}{2E7D32}
\definecolor{deltaneg}{HTML}{C62828}
\definecolor{deltazero}{HTML}{757575}

\newcommand{\sg}{\textsc{SGUID}\xspace}

\title{ Which Skill to Distill? \\
\sg \texttwemoji{squid}: Selecting a Compact Skill Bank for Model--Skill Co-Evolution}

\author{Yuhan Liu\textsuperscript{1}, Xiyao Ma\textsuperscript{2}, Zhongkai Sun\textsuperscript{2}, Xu Han\textsuperscript{2}, Chengyuan Ma\textsuperscript{2},\\
\bf Benjamin Z. Yao\textsuperscript{2}, Chenlei Guo\textsuperscript{2}\\[4pt]
{\normalfont New York University\textsuperscript{1}, Amazon\textsuperscript{2}}\\[2pt]
{\normalfont\texttt{yl13579@nyu.edu}}
}

\iclrfinalcopy 
\begin{document}

\maketitle
\lhead{Preprint. Under review.} 

\begin{abstract}
Skills, reusable procedural guidance added at inference, can substantially improve LLM downstream performance~\citep{li2026skillsbench}. 
Prior work retrieves skills from a bank by semantic relevance, then uses them as inference-time patches or for model distillation. The individual utility of each skill, however, is largely neglected.
We first show that, in on-policy distillation where skill-conditioned policies serve as teachers,  fewer than $25\%$ of retrieved skills provide useful distillation signals.
We then propose \sg, a method for selecting a compact subset of skills for distillation. \sg retains a skill only if it consistently yields effective learning signals during training. The selected skills are then distilled to produce a better model.
Our results show that \textit{not all} skills are worth distilling. Across four models from the Olmo and Qwen families, distilling 6 selected skills matches or exceeds full-bank distillation in mean avg@12 on three of the four models, and on all four after a second round that distills 3 newly selected skills, while the full banks are up to $11\times$ larger. 
Importantly, \sg supports stable model--skill co-evolution: after a distillation round, a new candidate bank is curated from the updated model's rollouts, and \sg selects which skills to internalize next. In the second round, this loop selects 3 new skills and improves \texttt{Qwen3-8B} from 64.3\% to 66.3\%. The selection step is essential for stability: on \texttt{Qwen3-4B}, naively updating the model with unfiltered skills degrades performance, including a $0.3$ percentage point drop on HMMT25, whereas \sg improves HMMT25 by $0.5$ points after the first round and $1.1$ points after the second. These results identify skill selection as the key mechanism for stable model--skill co-evolution.\footnote{We plan to release our code upon paper acceptance.}

\end{abstract}

\section{Introduction}


LLMs have demonstrated strong potential to advance challenging real-world domains, varying among reasoning~\citep{liu2024deepseek,georgiev2025mathematical,balunovic2026matharena,dekoninck2026beyond,Labiad2026BeyondRS}, coding~\citep{jimenez2024swe,yang2024swe,merrill2026terminal,ivison2026tmax,yang2026programbenchlanguagemodelsrebuild}, and multi-turn interactions~\citep{barres2025tau, lyu2026agenticqwen}, and more \citep{phan2025humanity,shao2025future}. Skills, reusable pieces of knowledge, serve as an intermediate layer between LLMs and downstream tasks. They assist the LLMs by augmenting the prompt with cognitive shortcuts~\citep{didolkar2025metacognitive}, task procedures, common mistake patterns, or structured packages of code and resources~\citep{anthropic2026skills,li2026skillsbench,shi2026skill1}. Such task-specific knowledge can be curated from human expertise, the model's own successes and failures, or a mixture of both \citep{ni2026trace2skill, yang2026skillopt, tang2026wikiskill}, and can be retrieved on request by task relevance \citep {li2025skillflow,xia2026skillrl}.
Skills are shown to be powerful tools that complement model strengths, bridging general-purpose capabilities and downstream expertise~\citep{li2026skillsbench,xie2026hybrid,didolkar2025metacognitive,ouyang2026reasoningbank,xia2026skillrl,ni2026trace2skill,yang2026skillopt}. 

 
Though skills patch model behavior, the improvements are often averaged over a bank of skills.  Yet the \textit{individual utility} is largely understudied. It is unclear if such gains are driven by all skills or only \textit{a fraction of} them. In SWE tasks, for example, \citet{han2026swe} find that only 7 out of 49 coding skills yield meaningful improvements, while many others have little effect or even degrade performance. This raises two coupled challenges for jointly improving the models and skill banks: how to develop and select non-trivial skills that enter the bank, and how to internalize skill-induced capabilities into the model.

Prior work addresses parts of this picture. Systems that maintain a skill bank select rather than accumulate, keeping only the skills whose usefulness is validated rather than assumed from relevance~\citep{fu2026davinci,shi2026skill1}. Meanwhile, on-policy self-distillation has been used to internalize skill-induced behavior into model weights, producing stronger LLMs that no longer require skill patches at inference~\citep{huang2026skill,wang2026skill}. However, it remains open how to identify the skills that are worth distilling, and whether internalizing those can support stable model--skill co-evolution.

To this end, we study skill selection and co-evolution via on-policy self-distillation, where skills serve as privileged context for teacher policies. We first find that across three Qwen models of varying sizes, fewer than $25\%$ of skills provide active learning signals. This suggests that most retrieved skills are not effective teachers for the current student. We then propose  \sg \texttwemoji{squid} (skill \textbf{S}election \textbf{GUI}ded by training \textbf{D}ynamics), a method that builds a compact skill bank by observing the training dynamics throughout the distillation. In detail, \sg takes on one distillation run with the whole skill bank, keeping the retrieved skill that produces a consistent and verifier-aligned distillation signal from early to late training. Those skills then form the compact bank used to update the student policy later. The updated model in turn samples rollouts from which a new candidate bank is curated. This whole procedure allows the skill bank and the model to evolve together.

We evaluate \sg on four models (\texttt{Olmo-3-7B}, \texttt{Qwen3-1.7B}, \texttt{Qwen3-4B}, and \texttt{Qwen3-8B}), with AIME24, AIME25, and HMMT25 benchmarks. \sg distills only 6 skills, yet matches or exceeds full-bank distillation in mean avg@12 on three of the four models, while the full banks are up to $11\times$ larger. On Qwen3-4B, using \sg instead of full-bank distillation improves performance from $64.2\%$ to $65.1\%$. Using \sg to distill 3 skills for a second round yields further gains:  \texttt{Qwen3-8B}'s performance rises from $64.3\%$ to $66.3\%$. Ablations further show that the improvements come from selecting useful and persistent skills rather than merely compressing the bank. Moreover, \sg is robust to the number of selected skills: compact banks with 4--12 skills remain competitive with, and sometimes outperform, the full bank. We also discuss whether \sg can be applied over an early training window instead of a full run, which lowers the cost of skill selection.

Our contributions are summarized as follows:
\begin{itemize}
    \item We propose \sg, a training-dynamics-based method for selecting compact skill banks for on-policy self-distillation. With only 6 skills, \sg matches or exceeds full-bank distillation in mean avg@12 on three of the four models, while the full banks are up to $11\times$ larger.
    \item We show that selected skills enable model--skill co-evolution. Using \sg for a second round improves \texttt{Qwen3-8B} mean avg@12 from $64.3\%$ to $66.3\%$, while co-evolution without skill selection hurts HMMT25 performance.
    \item We validate the selection signal through ablations on bank size, skill ranking, and the training-dynamics window. Compact banks with 4--12 skills remain competitive, top-ranked skills by our selection signals are most reliable, and training statistics from a 25-step run provide a promising but scale-dependent proxy for standard skill selection with \sg.
\end{itemize}

\section{Background}

\subsection{Problem Setting}
We study skill-guided self-distillation for verifiable tasks. Suppose we have a problem $x$ and a reference solution $y^\star$. A language model with parameters $\theta$ defines an autoregressive policy
$p_\theta(y\mid x)=\prod_{n=1}^{|y|}p_\theta(y_n\mid x,y_{<n})$.
 A skill bank $B_\theta$ can be curated from the current policy $\theta$'s rollouts, verified by the provided ground truth $y^\star$. Informally, for $\hat y\sim p_\theta(\cdot\mid x)$,
\begin{equation}
B_\theta=\operatorname{Curate}\!\left(\left\{(x,y^\star,\hat y):\hat y\sim p_\theta(\cdot\mid x)\right\}\right).
\label{eq:curate}
\end{equation}

Each skill $s\in B_\theta$ is a compact natural-language procedure, heuristic, or failure-mode description. Skills from the bank are retrieved to serve as training-time context for the teacher~\citep{huang2026skill, wang2026skill}, or as inference-time patches~\citep{yang2026skillopt,ni2026trace2skill}.

\subsection{On-Policy Self-Distillation}
On-Policy Self-Distillation (OPSD) turns privileged information into a dense training signal~\citep{zhao2026self,hubotter2026reinforcement}. Given $(x,y^\star)$, the teacher $P_T$, initialized from $\theta$, additionally conditions on $y^\star$, to judge the student $P_S$. This is calculated over responses sampled from the student $\hat y=(\hat y_1,\ldots,\hat y_{|\hat y|})\sim P_S(\cdot\mid x)$. Teacher and student therefore share same parameters and differ only in what they condition on. The relation holds at initialization; during training the teacher is either frozen at that copy or kept synchronized with the student, and we report which choice each method uses in Table~\ref{tab:hyperparams}. 
\begin{equation}
\begin{aligned}
P_S(y_n\mid x,\hat y_{<n})&=p_\theta(y_n\mid x,\hat y_{<n}),\\
P_T(y_n\mid x,y^\star,\hat y_{<n})&=p_\theta(y_n\mid x,y^\star,\hat y_{<n}).
\end{aligned}
\end{equation}

OPSD then minimizes the average next-token divergence (e.g., KL) between these two distributions over positions:
\begin{equation}
\mathcal{L}_{\mathrm{OPSD}}(\theta)
=\mathbb{E}_{(x,y^\star)}\mathbb{E}_{\hat y\sim P_S(\cdot\mid x)}
\left[D(P_T\|P_S)(\hat y\mid x,y^\star)\right].
\label{eq:opsd}
\end{equation}
The divergence $D$ can be further written in terms of the teacher--student log-probability gap. For a vocabulary token $v$, let
\begin{equation}
\Delta_{T,S}(v;x,y^\star,\hat y_{<n}):=\log P_T(v\mid x,y^\star,\hat y_{<n})-\log P_S(v\mid x,\hat y_{<n}).
\label{eq:log_gap}
\end{equation}
Then
\begin{equation}
D(P_T\|P_S)(\hat y\mid x,y^\star)=\frac{1}{|\hat y|}\sum_{n=1}^{|\hat y|}\mathbb{E}_{v\sim q_n}\left[\Delta_{T,S}(v;x,y^\star,\hat y_{<n})\right],
\end{equation}
where $q_n$ is a distribution over tokens at position $n$. \footnote{For the KL divergence $D_{\mathrm{KL}}(P_T\|P_S)$, $q_n=P_T(\cdot\mid x,y^\star,\hat y_{<n})$.}

\subsection{On-Policy Self-Distillation with Skills}
Researchers find that skill-guided teachers $p_\theta(\cdot\mid x,s)$ provide better learning signals than answer-guided ones $p_\theta(\cdot\mid x,y^\star)$ \citep{wang2026skill,huang2026skill}.  Here, a skill $s$ is retrieved based on the semantic distance between the problem $x$ and skill descriptions.

Both policies score the student rollout. For each sampled token $\hat y_n$, we can compute the teacher--student log-probability gap in \eqref{eq:log_gap}, with  the skill $s$ instead of $y^\star$: $\Delta_n(s;\hat y):=\Delta_{T,S}(\hat y_n;x,s,\hat y_{<n})$.
A positive gap means that the skill-conditioned teacher favors the sampled token more than the student. The plain support score
\begin{equation}
a(s,\hat y)=\frac{1}{|\hat y|}\sum_{n=1}^{|\hat y|}\Delta_n(s;\hat y)
\end{equation}
summarizes the teacher's stance on the rollout. In practice, SGSD~\citep{huang2026skill} uses a robust estimate $\tilde a(s,\hat y)$, obtained by masking irrelevant tokens and clipping extreme gaps before averaging.

The reference solution is used to verify the student's rollout in order to reverse the sign of the teacher signal when it disagrees with the verifier. Let $r(x,y^\star,\hat y)=+1$ if the verifier returns correct, and $-1$ otherwise. We write the resulting verifier-aligned teacher sign as $\operatorname{Verify}$:
\begin{equation}
\operatorname{Verify}(s,\hat y,y^\star)
=\operatorname{sign}\!\left(r(x,y^\star,\hat y)\,\tilde a(s,\hat y)\right).
\end{equation} 

A gate then filters out weak signals whose support-score magnitude $|\tilde a(s,\hat y)|$ is at most a confidence threshold $\kappa>0$:
\begin{equation}
\operatorname{Gate}_\kappa(\tilde a(s,\hat y))
=\mathbf{1}[|\tilde a(s,\hat y)|>\kappa],
\end{equation}
Using the same teacher--student divergence $D$ as in \eqref{eq:opsd}, now with the teacher conditioned on $s$, the single-skill form of the skill-guided self-distillation objective is
\begin{equation}
\mathcal{L}_{\mathrm{SGSD}}(\theta)
=\mathbb{E}_{(x,y^\star)}
\mathbb{E}_{\hat y\sim P_S(\cdot\mid x)}
\left[
\operatorname{Verify}(s,\hat y,y^\star)
\operatorname{Gate}_\kappa(\tilde a(s,\hat y))
D(P_T\|P_S)(\hat y\mid x,s)
\right].
\end{equation}
This formulation generalizes the original SGSD objective~\citep{huang2026skill}, which uses a particular bounded teacher--student discrepancy.

\section{Method}
\label{sec:method}

\begin{figure}[t]
\begin{center}
\includegraphics[width=\linewidth]{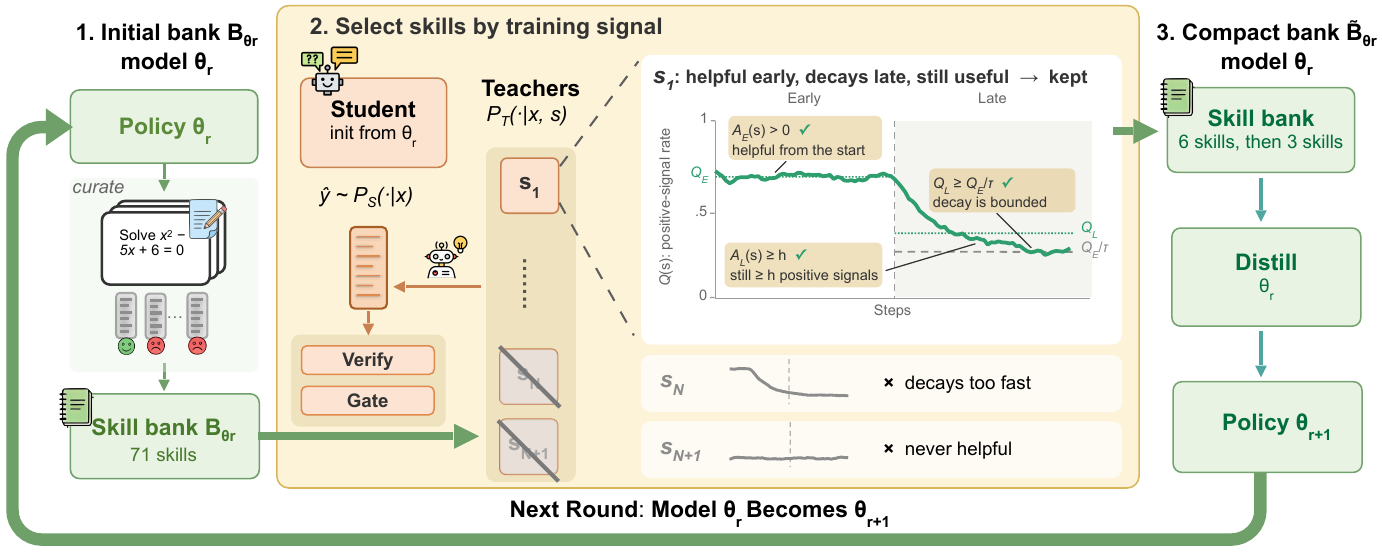}
\end{center}
\vspace{-5pt}
\caption{Overview of \sg at co-evolution round $r$. \textbf{(1) Curate:} the current policy $\theta_r$ rolls out on training problems, the rollouts are checked against the reference solutions $y^\star$ to curate a candidate skill bank $B_{\theta_r}$ (the bank size shown is that of \texttt{Qwen3-8B}). \textbf{(2) Select skills by training signal:} the student $P_S(\cdot\mid x)$, initialized from $\theta_r$, generates a rollout $\hat y$ that is scored by privileged teachers $P_T(\cdot\mid x,s_i)$, each conditioned on a retrieved skill $s_i$. Skills that contribute early, still produce at least $h$ positive events late, and whose contribution rate decreases by at most a factor of $\tau$ are kept; skills whose signal decays too fast, stays at zero, or that are never retrieved are discarded. \textbf{(3) Internalize:} restarting from $\theta_r$, the compact bank $\widetilde{B}_{\theta_r}$ with $|\widetilde{B}_{\theta_r}|=K\ll|B_{\theta_r}|$ is distilled into the model, producing $\theta_{r+1}$, which becomes the policy for the next round.}
\label{fig:overview}
\end{figure}


Figure~\ref{fig:overview} summarizes \sg as a three-stage model--skill co-evolution loop. Given a policy $\theta_{r}$ and its candidate skill bank $B_{\theta_r}$, our method selects a compact subset $\widetilde{B}_{\theta_r}\subseteq B_{\theta_r}$, retaining skills that consistently provide verifier-aligned teacher signals across training. We then distill this compact bank to obtain an updated policy $\theta_{r+1}$.

\begin{wrapfigure}{r}{0.4\linewidth}
\centering
\vspace{-15pt}
\includegraphics[width=\linewidth]{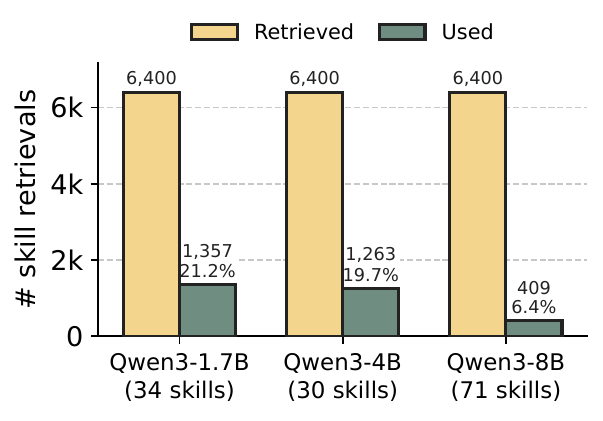}
\vspace{-20pt}
\caption{Skill retrieval does not guarantee active teacher signals. 
$ < 25\%$ of skills are active.}
\label{fig:retrieval_utility}
\vspace{-15pt}
\end{wrapfigure}
\paragraph{Motivation}
Retrieval surfaces potentially relevant skills, but it does not guarantee an active learning signal on a student rollout. Figure~\ref{fig:retrieval_utility} shows that, among 6,400 logged retrievals per model, only 21.2\% for Qwen3-1.7B, 19.7\% for Qwen3-4B, and 6.4\% for Qwen3-8B are actually used for distillation, while the rest are zeroed out by the gate.  This motivates us to ask whether \textit{a compact bank suffices},  including only persistently active skills. We hypothesize that persistently useful skills should be identified using both semantic relevance and the model's own training dynamics.

\paragraph{Skill Selection}
To identify skills that provide useful and persistent signals, we collect statistics during a training run with the full candidate skill bank $B_{\theta_r}$ of the current policy $\theta_r$. Let $\mathcal X=\{(x_i,y_i^\star)\}_{i=1}^{N}$ denote the training data. For each $(x,y^\star)\in\mathcal X$, the student samples a rollout $\hat y\sim P_S(\cdot\mid x)$. For each skill $s\in B_{\theta_r}$, we define its retrieval indicator as $I_{\mathrm{ret}}(s,x):=\mathbf{1}[s\text{ is retrieved for }x]$. When $I_{\mathrm{ret}}(s,x)=1$, we compute its signed training coefficient as
\begin{equation}
c(s;x,y^\star,\hat y)
=
\operatorname{Verify}(s,\hat y,y^\star)\,
\operatorname{Gate}_\kappa\!\left(\tilde a(s,\hat y)\right).
\label{eq:coef}
\end{equation}
We anticipate that a retrieved skill $s$ provides a useful contribution on an example when $c(s;x,y^\star,\hat y)>0$. A positive coefficient indicates that the skill-conditioned teacher provides an active signal aligned with the verifier.

We split the distillation into early and late phases, $p\in\{\mathrm E,\mathrm L\}$. For each skill $s$, let $N_p(s)$ denote its number of retrievals in phase $p$, and let $A_p(s)$ count the retrievals that produce a positive, verifier-aligned coefficient $c>0$. Whenever $N_p(s)>0$, its conditional contribution rate is $Q_p(s)=A_p(s)/N_p(s)$.

We retain skills that contribute in the early phase, produce at least $h$ positive signals in the late phase, and whose contribution rate decreases by at most a factor of $\tau>1$:
\begin{equation}
\mathcal F_{\theta_r}=
\left\{
s\in B_{\theta_r}:
A_{\mathrm E}(s)>0,\;
A_{\mathrm L}(s)\ge h,\;
Q_{\mathrm L}(s)\ge
\frac{Q_{\mathrm E}(s)}{\tau}
\right\}.
\label{eq:filter}
\end{equation}
Positive contribution counts imply positive retrieval counts, so both rates are well defined for every skill in $\mathcal F_{\theta_r}$. The late-phase count threshold avoids decisions based on sparse events, while the rate constraint removes skills whose useful signal fades during training. Finally, we select the $K$ candidates with the largest $A_{\mathrm L}(s)$, breaking ties by $Q_{\mathrm L}(s)$, to obtain $\widetilde{B}_{\theta_r}$.
 

\paragraph{Skill Internalization and Evolution}
After selecting $\widetilde{B}_{\theta_r}$, we restart from the same policy $\theta_r$ and train using only the compact bank. Let $\operatorname{Distill}(\theta,B)$ denote the parameters obtained by running the skill-conditioned gated distillation procedure from $\theta$ with bank $B$; the compact-bank update is
\begin{equation}
\theta_{r+1}
=
\operatorname{Distill}(\theta_r,\widetilde{B}_{\theta_r}).
\end{equation}

To continue the co-evolution, we repeat the same procedure with the
updated policy $\theta_{r+1}$. Specifically, we sample and verify rollouts
from $\theta_{r+1}$ and curate a new candidate bank $B_{\theta_{r+1}}$ using
\eqref{eq:curate}. Starting from $\theta_{r+1}$, we apply the same selection criterion based on training dynamics
to obtain a new compact bank $\widetilde{B}_{\theta_{r+1}}$, and distill only the newly selected skills:
\begin{equation}
\theta_{r+2}
=
\operatorname{Distill}(\theta_{r+1},\widetilde{B}_{\theta_{r+1}}).
\end{equation}


\begin{table}[!t]
\caption{Main results across three math reasoning benchmarks. We report the avg@12 accuracy for the best checkpoint according to the average performance on AIME24, AIME25, and HMMT25 across 200 training steps. \textbf{Bold} marks the best and \underline{underline} the second best. \# Distilled Skills is the number of skills the teacher can retrieve from; for SGSD we follow the original banks in \citet{huang2026skill}. R1 and R2 denote the first and second co-evolution rounds of \sg(ours). The skills selected for each model and round are listed in Appendix~\ref{app:skill_banks}.}

\label{tab:main_results}
\begin{center}
\begin{tabular}{llccccc}
\toprule
Model & Method & \# Distilled Skills & AIME24 & AIME25 & HMMT25 & Mean \\
\midrule
\multirow{6}{*}{Olmo-3-7B}
 & Base & -- & 53.1 & 39.4 & \textbf{26.9} & 39.8 \\
 & GRPO & -- & 53.6 & \underline{42.5} & 25.8 & 40.6 \\
 & OPSD & -- & 42.5 & 37.5 & 20.8 & 33.6 \\
 & SGSD & 38 & \underline{54.4} & 41.1 & 24.4 & 40.0 \\
\rowcolor{gray!15}
\cellcolor{white} & \sg, R1 & 6 & \textbf{55.0} & 42.2 & 25.6 & \underline{40.9} \\
\rowcolor{gray!15}
\cellcolor{white} & \sg, R2 & 3 & \textbf{55.0} & \textbf{42.8} & \underline{26.1} & \textbf{41.3} \\
\midrule
\multirow{6}{*}{Qwen3-1.7B}
 & Base & -- & 50.6 & 35.6 & 23.1 & 36.4 \\
 & GRPO & -- & 49.2 & 36.9 & 24.7 & 36.9 \\
 & OPSD & -- & \underline{56.4} & 41.7 & 27.5 & 41.9 \\
 & SGSD & 34 & 55.6 & \textbf{46.1} & 27.2 & \textbf{43.0} \\
\rowcolor{gray!15}
\cellcolor{white} & \sg, R1 & 6 & 55.3 & 43.1 & \underline{28.6} & \underline{42.3} \\
\rowcolor{gray!15}
\cellcolor{white} & \sg, R2 & 3 & \textbf{56.9} & \underline{43.3} & \textbf{28.9} & \textbf{43.0} \\
\midrule
\multirow{6}{*}{Qwen3-4B}
 & Base & -- & 73.9 & 65.8 & 46.7 & 62.1 \\
 & GRPO & -- & 76.1 & 66.7 & 45.3 & 62.7 \\
 & OPSD & -- & \textbf{77.8} & 68.9 & 46.1 & 64.3 \\
 & SGSD & 30 & 75.6 & 70.3 & 46.7 & 64.2 \\
\rowcolor{gray!15}
\cellcolor{white} & \sg, R1 & 6 & 76.7 & \textbf{71.4} & \underline{47.2} & \underline{65.1} \\
\rowcolor{gray!15}
\cellcolor{white} & \sg, R2 & 3 & \underline{76.9} & \underline{70.8} & \textbf{47.8} & \textbf{65.2} \\
\midrule
\multirow{6}{*}{Qwen3-8B}
 & Base & -- & 77.5 & 67.2 & 43.9 & 62.9 \\
 & GRPO & -- & 76.9 & 68.1 & 45.3 & 63.4 \\
 & OPSD & -- & \underline{78.1} & 69.7 & \textbf{48.1} & \underline{65.3} \\ 
 & SGSD & 71 & 77.2 & 67.2 & \underline{47.2} & 63.9 \\
\rowcolor{gray!15}
\cellcolor{white} & \sg, R1 & 6 & 75.3 & \underline{71.9} & 45.6 & 64.3 \\
\rowcolor{gray!15}
\cellcolor{white} & \sg, R2 & 3 & \textbf{78.3} & \textbf{73.9} & 46.7 & \textbf{66.3} \\
\bottomrule
\end{tabular}
\end{center}

\end{table}
\section{Experiments}

\subsection{Experimental Setup}
\label{sec:setup}

\paragraph{Models and Datasets} We train four models from two families, \texttt{Olmo-3-7B}, \texttt{Qwen3-1.7B}, \texttt{Qwen3-4B}, and \texttt{Qwen3-8B}, on DAPO-Math-17K~\citep{yu2026dapo}, following~\citet{wang2026beyond} and~\citet{huang2026skill}, and evaluate them on three math reasoning benchmarks: AIME24, AIME25, and HMMT25.

\paragraph{Baselines}
We compare against two types of methods: (1) \textbf{Learning w/ correct answers only}: GRPO~\citep{shao2024deepseekmath} compares the student's answer against the ground-truth answer to obtain a scalar reward. OPSD~\citep{zhao2026self} conditions the teacher policy on the ground truth to provide supervision for the student.
(2) \textbf{Learning w/ correct answers and skills}: SGSD~\citep{huang2026skill} conditions the teacher on a retrieved skill from a large bank to judge the student outputs. The teacher's supervision signal is then verified with the ground-truth answer. 
Following their setup, we train the models for 200 steps and report the checkpoint with the best avg@12.

\paragraph{Implementation Details}
We follow~\citet{huang2026skill} to build the initial skill bank. The candidate skills are generated by the student's own base model rather than by a stronger external model. During training, the thinking mode is disabled for the student and enabled for the teacher, following common practice~\citep{zhao2026self, huang2026skill}; at evaluation time, the thinking mode is enabled.  All experiments are conducted on 4 H200 GPUs with LoRA~\citep{hu2021lora}. The bank size is fixed in advance and shared by every model: we retain $K=6$ skills in round 1 and $K=3$ in round 2, motivated by training statistics showing that fewer skills continue to provide persistent signals after the first round of skill internalization. $K$ is never tuned per model, per benchmark, or against test performance, and the selection thresholds are set from training-time statistics alone; the ablation over $K$ in Table~\ref{tab:bank_size} is a post-hoc sensitivity check on a single model rather than the procedure used to choose the default. More experimental details are in Appendix \ref{app:exp_details}.

\subsection{Main findings}

\paragraph{A selected skill bank matches or outperforms the full bank.}
Table~\ref{tab:main_results} reports the main comparison on AIME24, AIME25, and HMMT25. GRPO yields only modest improvements over the base models, likely because trajectory-level rewards provide sparse supervision. On the Qwen3 models, OPSD and SGSD provide denser token-level supervision and yield larger gains than GRPO. Notably, \sg achieves comparable or better performance while distilling a much smaller skill bank. In round 1, \sg(R1) conditions the teacher on only 6 selected skills and already outperforms full-bank distillation (SGSD) across all three benchmarks on both \texttt{Olmo-3-7B} and \texttt{Qwen3-4B}; compared with SGSD, the mean scores improve from 40.0\% to 40.9\% and from 64.2\% to 65.1\%, respectively. On \texttt{Qwen3-4B}, \sg(R1) also improves the base model mean from 62.1\% to 65.1\% and outperforms OPSD by $0.8$ points. The advantage becomes clearer in the second round of co-evolution: with only 3 skills, \sg(R2) obtains the best mean score on \texttt{Olmo-3-7B}, \texttt{Qwen3-4B}, and \texttt{Qwen3-8B}, and matches SGSD on \texttt{Qwen3-1.7B}. These results suggest that distilling a small set of well-selected skills can outperform distilling the full retrieved bank of 30--71 skills, which contains up to $11\times$ more skills than the selected 6-skill bank.

\paragraph{\texttt{Qwen3-8B} benefits most from the second co-evolution round.}
SGSD outperforms OPSD on \texttt{Qwen3-1.7B} (43.0\% versus 41.9\%), indicating that retrieved skills compensate for a teacher that is still weak when given the answer alone. However, the gap closes on \texttt{Qwen3-4B} and reverses on \texttt{Qwen3-8B}, where SGSD retains only a $1.0$ point gain over the base model and falls behind answer-only OPSD by $1.4$ points, consistent with the observation in \citet{huang2026skill}. Once the answer-conditioned teacher is strong enough, skills retrieved from a large unfiltered bank appear to act as noise rather than guidance. In contrast, \sg does not suffer from this reversal: \sg(R2) outperforms OPSD at every scale. On \texttt{Qwen3-8B}, the second co-evolution round selects 3 new skills from the candidate bank curated from the updated model's rollouts and improves the mean from 64.3\% to 66.3\%, surpassing OPSD by $1.0$ points. This suggests that the degradation stems from \emph{which} skills the teacher is conditioned on rather than from skill conditioning itself.

\section{Ablation Studies and Discussion}

\begin{figure}[t]
\begin{center}
\includegraphics[width=0.65\linewidth]{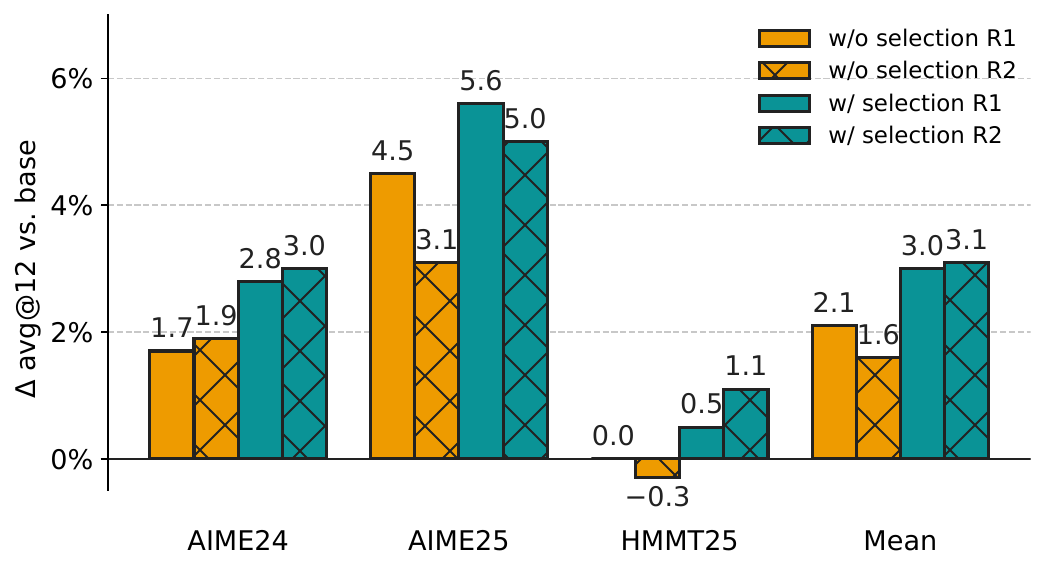}
\end{center}
\vspace{-10pt}
\caption{Two rounds of model--skill co-evolution on \texttt{Qwen3-4B}. Without selection (distilling the full bank), the mean gain shrinks from round one to round two; in contrast, \sg (w/ selection) improves the model steadily over rounds.}
\label{fig:co_evo_no_selection}
\end{figure}

\begin{table}[t]
\caption{\texttt{Qwen3-1.7B} and \texttt{Qwen3-4B} avg@12 when the compact bank is selected using training statistics from the full 200-step run (\sg (full)) or only the first 25 steps (\sg (25 steps)). The 25-step proxy matches the full run on 1.7B but is less reliable on 4B.}
\label{tab:cheap_selection}
\begin{center}
\begin{tabular}{llcccc}
\toprule
Model & Method & AIME24 & AIME25 & HMMT25 & Mean \\
\midrule
\multirow{3}{*}{Qwen3-1.7B}
 & Base & 50.6 & 35.6 & 23.1 & 36.4 \\ \cmidrule{2-6}
 & full, 200 steps & 55.3 & 43.1 & \textbf{28.6} & 42.3 \\
 & 25 steps & \textbf{56.1} & \textbf{43.9} & 27.5 & \textbf{42.5} \\
\midrule
\multirow{3}{*}{Qwen3-4B}
 & Base & 73.9 & 65.8 & 46.7 & 62.1 \\ \cmidrule{2-6}
 & full, 200 steps & \textbf{76.7} & \textbf{71.4} & \textbf{47.2} & \textbf{65.1} \\
 & 25 steps & 72.5 & 70.8 & 46.9 & 63.4 \\
\bottomrule
\end{tabular}
\end{center}
\end{table} 

\paragraph{Skill selection is necessary for stable co-evolution.}
We isolate the effect of skill selection by comparing co-evolution with the full skill bank against co-evolution with the selected bank (Figure~\ref{fig:co_evo_no_selection}). With the full bank, the mean avg@12 improvement over the base model drops from $2.1$ points in round 1 to $1.6$ points in round 2, mainly because the gain decreases on AIME25 from $4.5$ to $3.1$ points and turns negative on HMMT25 by $0.3$ points. In contrast, the selected bank gives stronger and more stable gains, improving the base model by $3.0$ points in round 1 and $3.1$ points in round 2. This suggests that skill selection is not only a compression step: it stabilizes model--skill co-evolution by filtering stale, redundant, or conflicting skills before they are reused in later distillation rounds.


\begin{wraptable}{l}{0.7\linewidth}
\vspace{-20pt}
\caption{How compact should the bank be? \texttt{Qwen3-4B} avg@12 as the number of retained skills $K$ varies. We use $K=6$ by default in the first round. The last row keeps the same budget but ranks skills by $A_{\mathrm L}(s)$ alone, dropping the persistence constraint $Q_{\mathrm L}(s)\ge Q_{\mathrm E}(s)/\tau$.}
\label{tab:bank_size}
\begin{center}
\setlength{\tabcolsep}{4pt}
\begin{tabular}{lcccc}
\toprule
Method & AIME24 & AIME25 & HMMT25 & Mean \\
\midrule
Base & 73.9 & 65.8 & 46.7 & 62.1 \\
OPSD & \textbf{77.8} & 68.9 & 46.1 & 64.3 \\
SGSD & 75.6 & 70.3 & 46.7 & 64.2 \\
\midrule
$K=2$ & 76.9 & 69.4 & 43.6 & 63.3 \\
$K=4$ & 75.6 & 70.8 & 46.9 & 64.4 \\
\rowcolor{gray!15}
$K=6$ & 76.7 & \textbf{71.4} & 47.2 & \textbf{65.1} \\
$K=12$ & 75.8 & 68.9 & \textbf{49.2} & 64.6 \\
\midrule
$K=6$, no persistence & 73.3 & 67.8 & 45.8 & 62.3 \\
\bottomrule
\end{tabular}
\end{center}
\vspace{-10pt}
\end{wraptable}
\paragraph{\sg is robust to the choice of bank size.}
We ablate the number of retained skills $K$ on \texttt{Qwen3-4B} (Table~\ref{tab:bank_size}). Banks of $K=4$, $6$, and $12$ all match or improve over SGSD while using far fewer skills, and the default $K=6$ is best on average, improving over the base model by $3.0$ points. Only the very small bank $K=2$ falls below SGSD, dropping on HMMT25, which suggests that too few skills lose useful coverage. Selection therefore does not require precise tuning of $K$, although an extremely small bank can hurt robustness. The persistence constraint, in contrast, is not optional: ranking by late-phase positive count alone and dropping the decay bound admits two skills whose contribution rate collapses between the early and late phase, and the mean falls to 62.3\%, $2.8$ points below \sg and barely above the base model. Selecting skills that are merely productive early is therefore not enough; they must keep contributing as the student improves.

\begin{wraptable}{r}{0.6\linewidth}
\vspace{-20pt}
\caption{\texttt{Qwen3-1.7B} avg@12 with $K=6$ held fixed, varying only which 6 skills are retained: the top-ranked, random (over 3 seeds for skill selection), or bottom-ranked (worst) 6 according to our selection signal. All skill selection baselines are trained for 100 steps. Bold marks the best value among the three $K=6$ variants; Base and SGSD are reference rows.}
\label{tab:mining_signal}
\begin{center}
\setlength{\tabcolsep}{4pt}
\begin{tabular}{lcccc}
\toprule
Method & AIME24 & AIME25 & HMMT25 & Mean \\
\midrule
Base & 50.6 & 35.6 & 23.1 & 36.4 \\
SGSD & 55.6 & 46.1 & 27.2 & 43.0 \\
\midrule
\rowcolor{gray!15}
top-6 & 55.3 & \textbf{43.1} & \textbf{28.6} & \textbf{42.3} \\
random-6 & 54.1{\scriptsize\,$\pm$0.8} & 42.6{\scriptsize\,$\pm$1.2} & 26.9{\scriptsize\,$\pm$0.9} & 41.2{\scriptsize\,$\pm$0.9} \\
worst-6  & \textbf{55.6} & 42.8 & 26.1 & 41.5 \\
\bottomrule
\end{tabular}
\end{center}
\vspace{-10pt}
\end{wraptable}
\paragraph{Top-ranked skills give the most reliable gains.}
We next fix $K=6$ and vary which 6 skills are retained on \texttt{Qwen3-1.7B} (Table~\ref{tab:mining_signal}), ranking the filtered candidates by their late-phase positive count $A_{\mathrm L}(s)$ and breaking ties by $Q_{\mathrm L}(s)$. The top-ranked bank is the best of the three variants on the mean and on AIME25 and HMMT25, coming within $0.7$ points of full-bank SGSD while conditioning the teacher on 6 skills instead of 34. Random and bottom-ranked banks are harder to tell apart and neither matches the top-ranked bank on the mean, so the gain comes from \emph{which} skills are kept rather than from a smaller bank alone.

\paragraph{Can we identify which skills to distill at lower cost?}
By default, \sg selects skills using training statistics collected over a full 200-step distillation run with the candidate bank. Table~\ref{tab:cheap_selection} asks whether this can instead be approximated from  the first 25 training steps. This approximation works well for \texttt{Qwen3-1.7B}, where the 25-step selected skills slightly improve the mean score over the full-run bank (42.5\% versus 42.3\%). However, it is less reliable for \texttt{Qwen3-4B}: the 25-step bank remains above the base model on average but trails the full-run bank by $1.7$ points (63.4\% versus 65.1\%). These results suggest that early training statistics contain useful information, but that reliably predicting the compact skill bank from a much earlier training stage remains an open direction for future work.

\section{Related Work}

\paragraph{Developing and maintaining the skill bank.}
External skill banks can be built through expert curation, community repositories, or agent experience. Although curated skills often improve agent performance, their effects vary across tasks and models, and compact, focused skills can outperform larger bundles~\citep{li2026skillsbench}. Accordingly, recent methods improve bank quality through large-scale retrieval~\citep{li2025skillflow}, trajectory consolidation~\citep{ni2026trace2skill}, validation-guided editing~\citep{yang2026skillopt,alzubi2026evoskill}, persistent knowledge accumulation~\citep{tang2026wikiskill}, or joint experience--skill memory~\citep{jiang2026xskill}. These methods primarily optimize external artifacts for inference with a fixed model. \sg instead closes the loop through the weights: the selected skills are internalized by distillation, and the updated model curates the next candidate bank, so selection is what carries improvement across rounds.

\paragraph{Skill internalization and model--skill co-evolution.}
On-policy distillation trains on student-generated trajectories with dense supervision signals from the teacher~\citep{agarwal2024policy}; self-distillation constructs a teacher from the same student policy, where the teacher is allowed to condition on extra privileged information such as ground-truth answers, solutions or environmental feedback~\citep{zhao2026self,hubotter2026reinforcement}. Recent work uses natural-language skills as additional privileged context for the teacher: Skill-SD distills trajectory-derived skills through a synchronized teacher, while SGSD validates each skill-conditioned teaching signal against verifier outcomes~\citep{wang2026skill,huang2026skill}. Complementary systems jointly evolve skill selection, utilization, and generation, or couple the skill bank with policy optimization, tools, persistent knowledge, verifiers, or self-play curricula~\citep{shi2026skill1,zhang2026co,wei2026skillsmith,zhang2026coevoskills,tang2026wikiskill,huang2026skillselfplay}. \sg builds on SGSD but shifts the focus from event-level distillation to bank-level curation: it aggregates verifier-aligned utility over training, selects a compact subset for a focused restart, and uses the updated model to seed the next skill bank.

\section{Conclusion}
Retrieval decides which skills a teacher \emph{sees}, not which ones actually teach. We find that fewer than 25\% of semantically retrieved skills produce active, verifier-aligned distillation signals, motivating \sg for compact-bank selection. \sg filters skills for persistent verifier-aligned contributions and then ranks the retained candidates by their late-phase positive contribution counts. Across four models, distilling 6 selected skills matches or exceeds full-bank distillation in mean avg@12 on three of the four models, while a second round with 3 newly selected skills does so on all four and improves \texttt{Qwen3-8B} from 64.3\% to 66.3\%. Selection also stabilizes co-evolution: on \texttt{Qwen3-4B}, selected banks preserve gains across rounds ($+3.0\rightarrow+3.1$ points), whereas unfiltered co-evolution decays ($+2.1\rightarrow+1.6$ points) and drops below the base model on HMMT25. These results show that skill selection is not merely bank compression, but a condition for stable model--skill co-evolution.

\section{Limitations}

We only test on single-turn math reasoning tasks. However, skills and distillation with skills via SFT have proven successful on various multi-turn agentic tasks~\citep{ouyang2026reasoningbank,xia2026skillrl,li2026skillsbench,ni2026trace2skill,yang2026skillopt}, and we therefore expect that our proposed model--skill co-evolution with skill selection can be generalized to multi-turn agentic tasks. We also focus on whether skills can serve as useful teachers in the training process, and we do not discuss whether such skills can serve as a useful inference-time patch. \citet{huang2026skill} find that augmenting base models with skills used in distillation could harm model performance at inference; we hypothesize that this is due to a train-test mismatch, during training, the student has thinking disabled and a maximum output length of 1,024 tokens. \citet{wu2026procedural}, however, find that conditioning on the math skills benefits math reasoning performance at inference time. We leave the co-evolution of skill sets that improve both training and inference to future work. 

\subsection*{AI use statement}

In this work, the authors only use generative AI to polish the text while the initial draft is written by humans. All AI-generated content is carefully verified by the authors. All research questions, experiments, and conclusions are developed and checked by the authors, without relying on AI assistants.

\subsection*{Reproducibility statement}

All models (\texttt{Olmo-3-7B}, \texttt{Qwen3-1.7B}, \texttt{Qwen3-4B}, and \texttt{Qwen3-8B}), the training set (DAPO-Math-17K), and the evaluation benchmarks (AIME24, AIME25, and HMMT25) are publicly available. The \sg selection rule is fully specified by Equations~\ref{eq:coef}--\ref{eq:filter} in Section~\ref{sec:method}, with the corresponding thresholds ($h$, $\tau$, the early/late split, and $K$) listed in Table~\ref{tab:hyperparams}. Training hyperparameters for all baselines and for \sg, including the LoRA configuration, batch sizes, and hardware, are given in Appendix~\ref{app:exp_details}; the evaluation protocol (avg@12 and thinking mode) is described in Section~\ref{sec:setup} and Appendix~\ref{app:exp_details}. The compact skill banks selected in each round for every model are listed verbatim in Appendix~\ref{app:skill_banks}, so that the distillation stage can be reproduced without re-running selection. We will release the source code, the skill-bank curation prompts, and the full candidate banks upon acceptance.

\bibliography{iclr2027_conference}
\bibliographystyle{iclr2027_conference}

\newpage
\appendix
\section{Experimental Details}
\label{app:exp_details}

\begin{table}[h]
\caption{Training hyperparameters for each method. LoRA is applied to \texttt{q\_proj}, \texttt{k\_proj}, \texttt{v\_proj}, \texttt{o\_proj}, \texttt{gate\_proj}, \texttt{up\_proj} and \texttt{down\_proj} in all runs. Dashes mark settings that do not apply.}
\label{tab:hyperparams}
\small
\begin{center}
\setlength{\tabcolsep}{4pt}
\begin{tabular}{lcccc}
\toprule
Hyperparameter & GRPO & OPSD & SGSD & \sg \\
\midrule
Learning rate & $5\times10^{-6}$ & $5\times10^{-6}$ & $5\times10^{-6}$ & $5\times10^{-6}$ \\
Completion length & 16000 & 1024 & 1024 & 1024 \\
Prompt length & 2048 & 20000 & 20000 & 20000 \\
Temperature & 1.2 & 1.1 & 1.1 & 1.1 \\
Top-$p$ / top-$k$ & -- & 0.95 / 20 & 0.95 / 20 & 0.95 / 20 \\
Generations per prompt & 8 & 1 & 1 & 1 \\
Teacher count per prompt & -- & 1 & 8 & 6 (R1) / 3 (R2) \\
Teacher update & -- & fixed & live (synchronized) & live (synchronized) \\
Skill retriever & -- & -- & \multicolumn{2}{c}{\texttt{Qwen3-Embedding-0.6B} (cosine)} \\
Gate threshold $\kappa$ & -- & -- & 0.05 & 0.05 \\
Support clip $\delta$ & -- & -- & 3.0 & 3.0 \\
Early/late split step & -- & -- & -- & 75 / 100 (R1) \\
Min late-phase positives $h$ & -- & -- & -- & 10 (R1) \\
Max rate decay $\tau$ & -- & -- & -- & 5 (R1) \\
Retained skills $K$ & -- & -- & -- & 6 (R1) / 3 (R2) \\
Per-device batch & 2 & 4 & 1 & 1 \\
Gradient accumulation & 4 & 2 & 4 & 4 \\
Effective batch & 32 & 32 & 16 & 16 \\
Training steps & 200 & 200 & 200 & 200 \\
Max grad norm & -- & 0.1 & 0.1 & 0.1 \\
LoRA rank $r$ & 64 & 64 & 64 & 64 \\
LoRA $\alpha$ & 128 & 128 & 128 & 128 \\
LoRA dropout & 0.05 & 0.05 & 0.05 & 0.05 \\
\bottomrule
\end{tabular}
\end{center}
\end{table}

\paragraph{Skill selection settings.}
Skills are retrieved by cosine similarity between the problem statement and the skill description. The compact-bank distillation run retrieves the $K$ selected skills and disables dynamic update. 


\newpage
\section{Compact Skill Banks}
\label{app:skill_banks}

Tables~\ref{tab:skills_qwen17}--\ref{tab:skills_olmo} list the compact skill banks that \sg selects for each model, in both co-evolution rounds. Each round curates its own candidate bank from the rollouts of the current policy, and skill identifiers are assigned within that bank. Identifiers are therefore local to a round: the same identifier in the round-1 and round-2 blocks of a table refers to two different skills.

\begin{table}[h]
\caption{Compact skill banks selected for \texttt{Qwen3-1.7B}.}
\label{tab:skills_qwen17}
\small
\setlength{\tabcolsep}{4pt}
\begin{center}
\begin{tabular}{lp{3.1cm}p{8.3cm}}
\toprule
ID & Title & Principle \\
\midrule
\multicolumn{3}{l}{\textit{Round 1 ($K=6$)}} \\
\texttt{gen\_007} & Solve triangle inequalities & Ensure sequences maintain triangle inequality conditions by enforcing each term $<$ sum of two preceding terms. \\
\texttt{gen\_017} & Balanced parity counting & When the number of even and odd elements are equal, the minimum number of swaps required is half the total number of elements. \\
\texttt{gen\_019} & Count with constraints & Calculate the number of valid configurations under specified constraints using combinatorial methods or factor pair analysis. \\
\texttt{gen\_021} & Integer factor pair analysis & Identify valid integer factor pairs for area or perimeter constraints to determine dimensions. \\
\texttt{gen\_023} & Count under constraints with digit analysis & Calculate the number of integers within a range that meet specific conditions, including multiples and digit patterns. \\
\texttt{gen\_025} & Pythagorean triple generation & Generate primitive Pythagorean triples using integer parameters $m > n$ and verify area-to-perimeter ratios. \\
\midrule
\multicolumn{3}{l}{\textit{Round 2 ($K=3$)}} \\
\texttt{gen\_007} & Use Modular Arithmetic & Leverage mathematical structures (e.g., prime factorizations, digit sums) and modular arithmetic to narrow candidate sets and validate constraints. \\
\texttt{gen\_009} & Compute 3D Distances & Calculate distances from points to lines in 3D using vector cross products: $|a \times b| / |b|$. \\
\texttt{gen\_026} & Number-Theoretic Analysis & Apply modular arithmetic, prime factorization, or parity checks to evaluate divisibility, congruence, or primality properties. \\
\bottomrule
\end{tabular}
\end{center}
\end{table}

\begin{table}[h]
\caption{Compact skill banks selected for \texttt{Qwen3-4B}.}
\label{tab:skills_qwen4}
\small
\setlength{\tabcolsep}{4pt}
\begin{center}
\begin{tabular}{lp{3.1cm}p{8.3cm}}
\toprule
ID & Title & Principle \\
\midrule
\multicolumn{3}{l}{\textit{Round 1 ($K=6$)}} \\
\texttt{gen\_002} & Simplify Expressions & Decompose complex expressions or ratios into simpler components using algebraic techniques, factoring, or number theory to reduce complexity. \\
\texttt{gen\_001} & Translate Constraints to Algebra & Convert geometric, number-theoretic, or combinatorial conditions into algebraic equations or expressions to isolate variables or derive relationships. \\
\texttt{gen\_006} & Analyze Symmetries & Adjust counts or configurations by dividing total possibilities by symmetry group order to account for equivalent arrangements. \\
\texttt{gen\_010} & Constraint Optimization & Maximize/minimize expressions or narrow down cases by analyzing variable relationships, parity, or integer properties under constraints. \\
\texttt{gen\_007} & Validate Solutions & Test candidate solutions against original constraints or edge cases to ensure consistency across scenarios. \\
\texttt{gen\_018} & Angle and Ratio Analysis & Apply trigonometric identities or vector proportionality to resolve relationships between angles, directions, or ratios. \\
\midrule
\multicolumn{3}{l}{\textit{Round 2 ($K=3$)}} \\
\texttt{gen\_006} & Number Theory and Optimization & Apply properties of primes, factorials, or modular arithmetic to simplify constraints or optimize number constructions. \\
\texttt{gen\_011} & Invariant and Recursive Reasoning & Identify invariants, thresholds, or recursive patterns to determine bounds, stability, or long-term behavior in sequences or systems. \\
\texttt{gen\_001} & Constraint and Probability Analysis & Translate constraints into equations or logical conditions, and apply combinatorial or probabilistic reasoning to evaluate outcomes under structured scenarios. \\
\bottomrule
\end{tabular}
\end{center}
\end{table}

\begin{table}[h]
\caption{Compact skill banks selected for \texttt{Qwen3-8B}.}
\label{tab:skills_qwen8}
\small
\setlength{\tabcolsep}{4pt}
\begin{center}
\begin{tabular}{lp{3.1cm}p{8.3cm}}
\toprule
ID & Title & Principle \\
\midrule
\multicolumn{3}{l}{\textit{Round 1 ($K=6$)}} \\
\texttt{gen\_045} & Geometric Constraint Modeling & Formulate algebraic equations from geometric relationships and properties. \\
\texttt{gen\_009} & Number Theory & Apply divisibility rules, modular arithmetic, and prime factorization to analyze number properties and solve equations. \\
\texttt{gen\_046} & Tangency Relationship Derivation & Use geometric perpendicularity properties to derive solution relationships. \\
\texttt{gen\_024} & Combinatorial Counting & Systematically count configurations by iterating through constraints or categorizing arrangements. \\
\texttt{gen\_039} & Integer Scaling Optimization & Minimize target quantities by testing minimal integer scaling parameters with common factors. \\
\texttt{gen\_042} & Domain Constraint Application & Limit analysis to feasible domains using function bounds and variable constraints. \\
\midrule
\multicolumn{3}{l}{\textit{Round 2 ($K=3$)}} \\
\texttt{gen\_004} & Case Analysis & Systematically count solutions by bounding search spaces or splitting into cases based on constraints. \\
\texttt{gen\_076} & Combinatorial Counting & Count configurations using permutations, combinations, or recursive decomposition of constraints. \\
\texttt{gen\_039} & Combinatorial Analysis & Systematically count configurations, permutations, and combinations with constraints using enumeration or algebraic techniques. \\
\bottomrule
\end{tabular}
\end{center}
\end{table}

\begin{table}[h]
\caption{Compact skill banks selected for \texttt{Olmo-3-7B}.}
\label{tab:skills_olmo}
\small
\setlength{\tabcolsep}{4pt}
\begin{center}
\begin{tabular}{lp{3.1cm}p{8.3cm}}
\toprule
ID & Title & Principle \\
\midrule
\multicolumn{3}{l}{\textit{Round 1 ($K=6$)}} \\
\texttt{gen\_027} & Diophantine Analysis with Parameterization & Analyze integer solutions by fixing parameters and applying Diophantine conditions. \\
\texttt{gen\_010} & Translate Verbal or Qualitative Problems to Mathematics & Convert non-technical or narrative problem statements into mathematical models or equations. \\
\texttt{gen\_002} & Simplify Expressions Using Algebra and Number Theory & Apply algebraic and number-theoretic techniques to reduce expression complexity. \\
\texttt{gen\_007} & Evaluate and Sum Series & Apply closed-form formulas to compute or analyze arithmetic, geometric, or power series. \\
\texttt{gen\_022} & Geometric Mean and Right Triangle Properties & Apply geometric mean properties of inscribed squares in right triangles to find areas or lengths. \\
\texttt{gen\_014} & Use Modular Arithmetic for Simplification and Validation & Simplify calculations using modular arithmetic for divisibility and large integers. \\
\midrule
\multicolumn{3}{l}{\textit{Round 2 ($K=3$)}} \\
\texttt{gen\_114} & Factorize and enumerate integer solutions & Use factorization of equations (e.g., quadratics) to enumerate valid integer parameters. \\
\texttt{gen\_009} & Planar triangulation and edge counting & For a convex $n$-vertex polygon, the maximum number of internal triangles in a triangulation is $T = 2n-4$. \\
\texttt{gen\_076} & Apply divisibility rules to transform sum constraints & Convert divisibility requirements into sum conditions (e.g., digit sums for divisibility by 3 or 9). \\
\bottomrule
\end{tabular}
\end{center}
\end{table}

\end{document}

%% file: math_commands.tex
\usepackage{amsmath,amsfonts,bm}

\def\eqref#1{equation~\ref{#1}}

\def\1{\bm{1}}

\DeclareMathAlphabet{\mathsfit}{\encodingdefault}{\sfdefault}{m}{sl}
\SetMathAlphabet{\mathsfit}{bold}{\encodingdefault}{\sfdefault}{bx}{n}

